\documentclass[11pt,a4paper]{article}
\usepackage[hyperref]{acl2020}
\usepackage{times}
\usepackage{latexsym}

\usepackage{graphicx}
\usepackage{amsmath}
\usepackage[ruled]{algorithm2e}
\SetAlFnt{\small}
\SetAlCapFnt{\large}
\SetAlCapNameFnt{\large}
\usepackage{xcolor}

\usepackage{url}

\usepackage{microtype}

\aclfinalcopy % Uncomment this line for the final submission
\def\HiLi{\leavevmode\rlap{\hbox to \hsize{\color{yellow!50}\leaders\hrule height .8\baselineskip depth .5ex\hfill}}}

\def\HiLi{\leavevmode\rlap{\hbox to \hsize{\color{yellow!50}\leaders\hrule height .8\baselineskip depth .5ex\hfill}}}
	\definecolor{azure(web)(azuremist)}{rgb}{0.94, 1.0, 1.0}
	\definecolor{babyblue}{rgb}{0.54, 0.81, 0.94}

\title{Active Learning for Coreference Resolution using Discrete Annotation}

\author{Belinda Z. Li$^{\dagger}\thanks{*Work done while at the University of Washington.}$ \quad \quad
Gabriel Stanovsky$^{\spadesuit\diamondsuit}$ \quad \quad
   Luke Zettlemoyer$^{\spadesuit\dagger}$ \quad \quad \\
   $^\spadesuit$
  University of Washington \quad \quad
  $^\diamondsuit$Allen Institute for AI \quad \quad
  $^\dagger$Facebook 
 \\ 
 {\tt belindali@fb.com}\\
  {\tt \{gabis,lsz\}@cs.washington.edu}
}

\date{}

\begin{document}
\maketitle
\begin{abstract}
  We improve upon
pairwise annotation for active learning in coreference resolution, by asking annotators to identify mention antecedents if a presented mention pair is deemed not coreferent.
This simple modification, when combined with a novel mention clustering algorithm for selecting which examples to label, is much more efficient in terms of the performance obtained per annotation budget. In experiments with existing benchmark coreference datasets, we show that the signal from this additional question leads to significant performance gains per human-annotation hour.
Future work can use our annotation protocol to effectively develop coreference models for new domains.
Our code is publicly available.\footnote{\url{https://github.com/belindal/discrete-active-learning-coref}}
\end{abstract}

\section{Introduction}

% \begin{itemize}
Coreference resolution is the task of resolving anaphoric expressions to their antecedents (see Figure~\ref{fig:example}). It is often required in downstream applications such as question answering \cite{Dasigi_2019} or machine translation \cite{Stanovsky2019ACL}.
Exhaustively annotating coreference is an expensive process as it requires 
tracking coreference chains across long passages of text.
In news stories, for example, important entities may be referenced many paragraphs after their  introduction.
%This puts a high barrier for annotating coreference data on larger datasets and new domains. 

% fully-labelled datasets are often difficult to obtain...
% However, coreference resolution systems often benefit from learning from labelled data... (because???)

\emph{Active learning} is a technique which aims to reduce costs by annotating samples which will be most beneficial for the learning process, rather than fully labeling a large fixed training set.
Active learning consists of two components: (1) a task-specific learning algorithm, and (2) an iterative sample selection algorithm, which examines the performance of the model trained at the previous iteration and selects samples to add to the annotated training set. This method has proven successful for various tasks in low-resource domains
% This method has proven successful in various tasks and practical applications where data is scarce, e.g., in low-resource languages or domains 
\cite{garrette-baldridge-2013-learning,kholghi2015active,7472693,7953171}.%,liu-etal-2018-learning-actively}.

\newcite{Sachan:2015:ALA:2832415.2832432} showed that active learning can be employed for the coreference resolution task. They used gold data to simulate pairwise human-annotations, where two entity mentions are annotated as either coreferring or not (see first question in Figure \ref{fig:example}).
%Samples to annotate were selected based on ...
%They estimated that reasonable performance can be obtained with less than 50 hours of human annotation time.%, using a clustering coreference resolution model and various sampling techniques.

\begin{figure}
    \centering
    \small
    \begin{tabular}{|p{2.85in}|} \hline
        % Late pictures into us during the commercial break. A volcano in Mexico that's just started spewing molten rock, known to locals as Po-po. If you've landed in Mexico City, you've seen \colorbox{green}{it}. About 40 miles outside town. When \colorbox{gray}{it} started showering ash a few days ago, authorities there started urging tens of thousands of people nearby to leave. A lot of people however still refusing. The volcano is almost 18,000 feet high. Molten rock is spewing, is reportedly setting trees afire, in the area. \\
        % \textcolor{red}{Are these two coreferent? y/[n]:} \textcolor{blue}{y} \\ \hline \hline
       A volcano in Mexico, known to \colorbox{babyblue}{locals} as \colorbox{yellow}{Po-po}, just started spewing molten rock. \\%If you've landed in Mexico City, you've seen it, just outside town. When it started showering ash a few days ago, authorities there started urging tens of thousands of people nearby to leave. The volcano is almost 18,000 feet high. Molten rock is spewing, is reportedly setting trees afire. \\
        \textcolor{red}{Are the two mentions coreferent?} \textcolor{blue}{No} \\
        \textcolor{red}{What is the \emph{first} appearance of the entity that the yellow-highlighted text refers to?} \textcolor{blue}{A volcano in Mexico} \\ \hline
    \end{tabular}
    \caption{Discrete annotation. The annotator is shown the document, a span (yellow), and the span's predicted antecedent (blue). 
    In case the answer to the coreference question is negative (i.e., the spans are not coreferring), we present a follow-up question (``what is the \emph{first} appearance of the entity?"), providing additional cost-effective signal. Our annotation interface can be seen in Figure~\ref{fig:interface} in the Appendix.}
    % \\
    % Top: scenario where user answers ``yes" to the initial question, prompting the next example. \\
    % Bottom: scenario where user answers ``no", prompting follow-up question.
    \label{fig:example}
\end{figure}

In this paper, we propose two improvements to active learning for coreference resolution.
First, we introduce the notion of \emph{discrete annotation} (Section~\ref{sec:discrete}), which augments pairwise annotation by introducing a simple additional question: if the user deems the two mentions non-coreferring, they are asked to mark the first occurrence of one of the mentions (see second question in Figure \ref{fig:example}). We show that this simple addition has several positive implications. %, significantly improving the quality of the models obtained per human annotation time.
The feedback is relatively easy for annotators to give, and provides meaningful signal which dramatically reduces the number of annotations needed to fully label a document. 
% The number of discrete questions that can be asked per document is far less than the number of pairwise questions. 

 Second, we introduce \emph{mention clustering} (Section~\ref{sec:active}). 
%  Discrete selection allows us to introduce an improvement to instance selection.  
When selecting the next mention to label, we take into account aggregate model predictions for all antecedents which belong to the same cluster. This avoids repeated labeling that would come with separately verifying every mention pair within the same cluster, as done in previous methods. 
%Thus, while in unclustered selection, the classes across which we compute selection metrics are candidate antecedent, in mention-clustered selection, the classes across which we compute selection metrics are candidate \emph{clusters}.

%During each iteration, we can sum the model 
%a --(word that is not ``augmentation")-- to mention selection for active learning annotation that accounts for . 

%a task-specific augmentation to selection, where instead of computing uncertainty metrics over individual antecedents or mentions, we first aggregate scores for antecedents of the same coreference cluster before computing these metrics. %We apply clustering to three distinct selection algorithms for active learning and show that it can bring concrete performance improvements to our system.

We conduct experiments across several sample selection algorithms using existing gold data for user labels and show that both of our contributions significantly improve performance on the CoNLL-2012 dataset \cite{pradhan2012conll}.
Overall, our active learning method presents a superior alternative to pairwise annotation for coreference resolution, achieving better performing models for a given annotation budget.

% We further improve performance by introducing \emph{clustering} to these selection algorithms, where we first aggregate model scores for antecedents in the same coreference cluster before applying the selectors. 
% We provide details into our clustering technique in section \ref{sec:selectors}.
%\gabis{Say something about the clustering technique for sampling?}

\section{Background}

Our work relies on two main components:
a coreference resolution model and a sample selection algorithm.%, which selects samples to label and add to the model's training set at each step.
%In this section, we will cover the most relevant works for each of these components.

\paragraph{Coreference resolution model}
% Coreference resolution has been an active area of research. %, with the main model types being mention-pair classifiers \cite{ng-cardie-2002-identifying,bengtson-roth-2008-understanding}, entity-level models \cite{haghighi-klein-2010-coreference,clark-manning-2015-entity,clark-manning-2016-improving,wiseman-etal-2016-learning}, latent-tree models \cite{fernandes-etal-2012-latent,bjorkelund-kuhn-2014-learning,martschat-strube-2015-latent},
%and ranking models 
% One type of model that has been proposed is the \emph{ranking} model, which given a mention or span $i$, ranks candidate antecedents by how likely they are to be antecedents of $i$. \cite{Lee2017EndtoendNC,clark-manning-2016-deep,durrett-klein-2013-easy,wiseman-etal-2015-learning}. 
We use the span ranking model introduced by \newcite{Lee2017EndtoendNC}, and later implemented in AllenNLP framework \cite{Gardner2018AllenNLPAD}.
This model computes span embeddings for all possible spans $i$ in a document, and uses them to compute a probability distribution $P(y=\text{ant}(i))$ over the set of all candidate antecedents $\mathcal{Y}(i) = \{K$ previous mentions in the document$\}\cup \{\epsilon\}$, where $\epsilon$ is a dummy antecedent signifying that span $i$ has no antecedent. This model does not require additional resources, such as syntactic dependencies or named entity recognition, and is thus well-suited for active learning scenarios for low-resource domains.

% They assigned to each span $i$ an antecedent that is either one of the up to $K$ spans to its left, or a dummy antecedent. Assigning the dummy antecedent signifies $i$ has no antecedent.
% They assigned each span $i$ an antecedent $A(i)$ from the set $\mathcal{Y}(i)$, whereby $A(i) = \max_{y\in \mathcal{Y}(i)}(P(y=\text{ant}(i)))$, and $\mathcal{Y}(i) = \{K$ previous mentions in the document$\}\cup \{\epsilon\}$. $\epsilon$ is a dummy antecedent signifying that span $i$ has no antecedent.

%We use the AllenNLP \cite{Gardner2018AllenNLPAD} implementation of the  model for co-reference. \newcite{Lee2017EndtoendNC}'s model works 

\paragraph{Sample selection algorithm}
Previous approaches for the annotation of coreference resolution 
have used mostly \emph{pairwise selection}, where pairs of mentions are 
shown to a human annotator who marks whether they are co-referring \cite{Gasperin:2009:ALA:1564131.1564133,laws-etal-2012-active, zhao-ng-2014-domain, Sachan:2015:ALA:2832415.2832432}.
To incorporate these binary annotations into % an active learning framework where documents are only partially labeled at each iteration, 
their clustering coreference model, \newcite{Sachan:2015:ALA:2832415.2832432} introduced
the notion of \emph{must-link} and \emph{cannot-link} penalties, which we describe and extend in Section~\ref{sec:active}.
%These are incurred to the model's loss function during training, whenever its prediction violates any of the available human gold annotations.
%In Section \ref{sec:active} we adapt this notion %\newcite{Sachan:2015:ALA:2832415.2832432}'s approach
%for our purposes.%, and formally define the must-link and cannot-link penalties we use in our model.

%Active learning for coreference resolution has mostly used either document-wise \cite{Miller:2012:ALC:2391123.2391133} or pairwise \cite{Gasperin:2009:ALA:1564131.1564133, laws-etal-2012-active, zhao-ng-2014-domain, Sachan:2015:ALA:2832415.2832432} annotation. In document-wise annotation, entire documents are selected and annotated by active learning, while in pairwise annotation,  \newcite{Sachan:2015:ALA:2832415.2832432} showed that pairwise annotation is more efficient than document-wise annotation. 

%We also use a timing experiment to prove that discrete annotation is more effective than pairwise.
%\newcite{laws-etal-2012-active} also uses a variant of pairwise annotation, N-pooling, whereby a neighborhood of mention pairs is selected and annotated at once.
%In this paper, we introduce a novel annotation scheme called \textit{discrete annotation}.
% As far as we know, this work is the first active learning for coreference resolution work to use a ranking model. 
% whereby a pair is added to must-link upon positive judgment of coreference, and is otherwise added to cannot-link.
% We borrow this idea, but with some adaptions, which we introduce in \ref{sec:links}.

\section{Discrete Annotation}
\label{sec:discrete}
%In this section we formally define \emph{discrete annotation} for coreference resolution.
%, which extends upon the pairwise annotation scheme.

% \paragraph{Formal Definition.}
In {\em discrete annotation}, as exemplified in Figure~\ref{fig:example}, we present the annotator with a document where the least certain span $i$ (``Po-po", in the example) and $i$'s model-predicted antecedent, $A(i)$ (``locals"),  % (defined as $\max_{y\in \mathcal{Y}(i)}(P(y=\text{ant}(i)))$)
are highlighted.
%and two highlighted spans $m$ and $a$, where $m$ is a mention and $a\in\mathcal{Y}(m)$ is a candidate antecedent for that mention,
%\gabis{This assumes that $a$ and $m$ are asymmetric? one is an antecedent and the other a mention? If so, this is a deviation from the symmetric pairwise, no?} \belinda{yes they are asymmetric} 
Similarly to pairwise annotation, annotators are first asked whether $i$ and $A(i)$ are coreferent. If they answer positively, we move on to the next sample. Otherwise, we deviate from pairwise sampling and ask the annotator to mark the antecedent for $i$ (``A volcano in Mexico") as the \emph{follow-up} question.\footnote{For consistency, we ask annotators to select the \textit{first} antecedent of $i$ in the document.}
The annotator can abstain from answering the follow-up question in case $i$ is not a valid mention or if it does not have an antecedent in the document. See Figure~\ref{fig:interface} in the Appendix for more example annotations.

% Thus, the user simply needs to scan the beginning of the document up to $m$, for possible other mentions of $m$.

% An important distinction between discrete and pairwise annotation is that in discrete annotation, we select the next most uncertain \textit{mention}, rather than the next most uncertain \textit{pair} of mentions, to annotate. Therefore, the initial pair that we present in discrete annotation is that mention $m$ and its most likely, non-$\epsilon$ antecedent $a\in\mathcal{Y}(m)\backslash\epsilon$.

% \paragraph{Motivation.}
In Section~\ref{sec:evaluation}, we show that discrete annotation is superior to the classic pairwise annotation in several aspects. First, it makes better use of human annotation time, as often an annotator needs to resolve the antecedent of the presented mention to answer the first question. For example, identifying that ``Po-po" refers to the volcano, and not the locals.
Second, we find that discrete annotation is a better fit for mention ranking models~\cite{Lee2017EndtoendNC}, which assign the most-likely antecedent to each mention, just as an annotator does in discrete annotation.
% In Section~\ref{sec:evaluation}, we show that discrete annotation indeed significantly improves model performance per human-annotation time.
% Furthermore, discrete annotation is better suited for a ranking model~\cite{Lee2017EndtoendNC}, which assigns the most-likely antecedent to each mention, just as an annotator does in discrete annotation.
% may already know what is the correc
% that discrete annotation makes better use of human annotation budget. since the annotator has already identified the entity within the document in order to answer, we hypothesize that the follow-up question would take less additional time to achieve the same performance improvement as additional pairwise questions.

% takes relatively little additional cognitive effort over pairwise annotation, while providing meaningful additional signal. The initial question is identical to pairwise annotation, thus discrete annotation only requires additional effort if the initial question was answered negatively.
% % Note we preserve the initial question as it saves us time in the case it is answered positively.
% In this case, since the annotator has already identified the entity within the document in order to answer, we hypothesize that the follow-up question would take less additional time to achieve the same performance improvement as additional pairwise questions. %not take much additional time. % over the initial question.

\section{Mention Clustering}
%\section{Choosing The Next Mention To Query}
%\section{Active Learning}
\label{sec:active}
%We experiment with three selection techniques to choose the next mention to query for discrete annotation.\footnote{See Algorithm \ref{alg:train} in the Appendix for full details on our discrete annotation active learning training loop.} We introduce the idea of clustering before applying popular active learning selectors like entropy or query-by-committee \cite{settles2010active}.
We experiment with three selection techniques by applying popular active learning selectors like entropy or query-by-committee \cite{settles2010active} to \emph{clusters} of spans. %\footnote{See Algorithm \ref{alg:train} in the Appendix for full details on our discrete annotation active learning training loop.}
Because our model outputs antecedent probabilities and predictions, we would like to %it makes sense to
aggregate these outputs, such that we have only one probability per mention cluster rather than one per antecedent. %This allows us to avoid treating differently antecedents which the model already predicted to belong in the same cluster.
We motivate this with an example: 
%To illustrate why we use clustered probabilities rather than antecedent probabilities, consider the following example:
suppose span $i$'s top two most likely antecedents are $y_1$ and $y_2$. In scenario 1, $y_1$ and $y_2$ are predicted to be clustered together, and in scenario 2, they are predicted to be clustered apart. Span $i$ should have a ``higher certainty"  in scenario 1 (and thus be less likely to be picked by active learning), because its two most likely antecedents both imply the same clustering, whereas in scenario 2, picking $y_1$ vs. $y_2$ results in a different downstream clustering. Thus, rather than simply using the raw probability $i$ refers to a particular antecedents, we use the probability \textit{$i$ belongs to a certain cluster}. This implies modelling $y_1$ and $y_2$ ``jointly" in scenario 1, and separately in scenario 2.

Formally, we compute the probability that a span $i$ belongs in a cluster $C$ by summing $P(\text{ant}(i) = y)$ for all $y$ that belong in some cluster $C$, since $i$ having an antecedent in a cluster necessarily also implies $i$ is also in that cluster. This allows us to convert the predicted antecedent probabilities to in-cluster probabilities:
\begin{align}
    \label{eq:clustered_probabilities}
    P(i\in C)= \sum_{y\in C\cap\mathcal{Y}(i)} P(\text{ant}(i) = y)
\end{align}

Similarly, for
query-by-committee, we aggregate predictions such that we have one vote per cluster rather than one vote per antecedent:
\begin{align}
    \label{eq:clustered_votes}
    V(i\in C) = \sum_{y\in C\cap\mathcal{Y}(i)} V(A(i) = y)
\end{align}
where $V(A(i) = y)\in\{0,1,\cdots, \mathcal{M}\}$ refers to the number of models that voted $y$ to be the antecedent of $i$. %To evaluate $y\in C\cap\mathcal{Y}(i)$, we use cluster information from the \textit{ensemble} model, plus any active learning labels.
%This allows us to avoid treating antecedents which the model predicted to belong in the same cluster differently.% Linking a mention to $a_1$ rather than $a_2$ makes no difference in the final clustering if $a_1$ and $a_2$ were predicted to be in the same cluster, so it doesn't make sense to treat them differently.

The cluster information ($y\in C\cap\mathcal{Y}(i)$) we use in Equations \ref{eq:clustered_probabilities} and \ref{eq:clustered_votes} is computed from a combination of model-predicted labels and labels queried through active learning. Antecedents which were not predicted to be in clusters are treated as singleton clusters.

%Following~\newcite{Sachan:2015:ALA:2832415.2832432}, we enforce user annotations during the selection process by keeping track of 
%\newcite{Sachan:2015:ALA:2832415.2832432} and keep track of
%\emph{must-link} (ML) and \emph{cannot-link} (CL) relations between mentions in accordance with human judgments.
Additionally, to respect user annotations during the selection process, we must keep track of all prior annotations. To do this, we use the concept of \emph{must-link} (ML; if two mentions are judged coreferent) and \emph{cannot-link} (CL; if two mentions are judged non-coreferent) relations between mentions introduced by \newcite{Sachan:2015:ALA:2832415.2832432}, and adapt it for our purposes.
Specifically, in our discrete setting, we build the links as follows: if the user deems the  pair coreferent, it is added to ML. Otherwise, it is added to CL, while the user-corrected pair (from the second question) is always added to ML.

% Link relations were originally an idea introduced by \newcite{Sachan:2015:ALA:2832415.2832432} as a means to incorporate binary user judgments into their global clustering model.\footnote{They did this by incurring penalties on their objective function. We incur penalties as well (though not for this purpose), see section 6.2 in appendix for more details on how we adapt the penalties for our specific loss function.} 
% However, beyond \newcite{Sachan:2015:ALA:2832415.2832432},
In addition, we use these links to guide how we \textit{select} for the next mention to query. For example, if a CL relation exists between spans $m_1$ and $m_2$, we will be less likely to query for $m_1$, since we are slightly more certain on what $m_1$'s antecedent should be (not $m_2$). %Similarly, if an ML relation already exists out of $m_1$, then we know we do not need to query for it.
Formally, we revise probabilities and votes $P(i\in C)$ and $V(i\in C)$ %(which are used to compute uncertainty scores for our 3 selectors)
in accordance to our link relations, which affects the selector uncertainty scores.\footnote{See Section~\ref{sec:model_appendix}  in the appendix for more details.}

% On the other hand, if a ML relation existed between $m_1$ and $m_2$, we set $P(ant(m_1) = m_2) = 1$ and thus will be less likely to query for $m_1$ under uncertainty sampling.
%We adapt these relations to work in our discrete setting:
%We incur the following penalties on our loss function:
%\begin{align}
%    P_{ML} = -\text{log}\sum_{\substack{(i, y_i)\in \text{ML}\land y_i\in\mathcal{Y}(i)\\ \land C(i)\neq C(y_i)}}P(y_i=\text{ant}(i)) \\
%    P_{CL} = \text{log}\sum_{\substack{(i, y_i)\in \text{CL}\land y_i\in\mathcal{Y}(i)\\ \land y_i=A(i)}}P(y_i=\text{ant}(i))
%\end{align}
%$C(i), C(y_i)$ signify the latent model-predicted clusters of $i$ and $y_i$.% Note $P_{ML}$ increases as the probability that two must-link mentions are linked together decreases, while $P_{CL}$ increases as the probability two cannot-link mentions are linked together increases.
% Note that in the summation for $P_{ML}$, we check the cluster equality of $i$ and $y_i$. While it would be simpler to check antecedent equality due to model design (which is what we do for the $P_{CL}$), we \textit{must} check cluster equality in the must-link case to avoid incorrectly penalize the scenario where the model clusters two mentions $i$ with $y_i$ indirectly through a sequence of links (such that $y_i\neq A(i)$ but $C(i) = C(y_i)$).  \gabis{Is this a special consideration in our case, if not, I think we can leave this detail out.}

Finally, following \cite{Sachan:2015:ALA:2832415.2832432}, we impose transitivity constraints, which allow us to model links beyond what has been explicitly pointed out during annotation:
%To ensure annotation consistency and 
%explicate links implied by human annotators, we also adopt \newcite{Sachan:2015:ALA:2832415.2832432} transitivity constraints:
%Moreover, we introduce an algorithm to incrementally compute transitive closures (algorithm \ref{alg:inc_closures}):
\begin{align}\label{eq:closure1}
ML(m_i, m_j)\land ML(m_j, m_k)\to ML(m_i, m_k) \\
CL(m_i, m_j)\land ML(m_i, m_k)\to CL(m_j, m_k) \label{eq:closure2}
 \end{align}
% and also modify the link penalties to suit our model's loss function.

% \paragraph{Incremental Link Closures.}
%Although we perform multi-instance selection, we update the above-mentioned link closures after every annotation instance, rather than at the end of a round. 
However, recomputing these closures after each active learning iteration can be extremely inefficient. Instead, we build up the closure incrementally by adding only the minimum number of necessary links to maintain the closure every time a new link is added.%\footnote{See Algorithm \ref{alg:inc_closures} and section \ref{sec:model_appendix} in the Appendix.}

% Computing closures is meant as a way of deducing links which the user may not have explicitly pointed out during annotation. By choosing to compute closures after each user annotation rather than at the end of a round, we can use this information when we're in the middle of a round to inform which annotation to ask for next. We describe this process in detail in section \ref{sec:selectors}.

% \paragraph{Link Penalties.}
%\gabis{Is this taken from Sachan?}
%\belinda{The idea of adding link penalties to loss fxn are, but these specific formulas are unique}
% Following \newcite{Sachan:2015:ALA:2832415.2832432}, we add penalties $P_{ML}$ and $P_{CL}$ to the loss function:
%, but we tailor the penalties to our model's loss function: %Let $l_{ori}$ represent the original loss function used by \newcite{Lee2017EndtoendNC}'s model.
% The original loss function used by \newcite{Lee2017EndtoendNC}'s model was
%     $$l_{ori} = -\text{log}\prod_{i=1}^N\sum_{\hat{y}\in \mathcal{Y}(i)\cap \text{GOLD}(i)}P(\hat{y}=\text{ant}(i))$$
%We use the modified loss function $l = l_{ori} + P_{ML} + P_{CL}$, where penalties $P_{ML}$ and $P_{CL}$ are defined as follows:

We experiment with the following clustered selection techniques:
\label{sec:selectors}
\paragraph{Clustered entropy}
We compute entropy over cluster probabilities and select the mention with the highest \textit{clustered} entropy:
\begin{align}
    E(i) = -\sum_{C\in\text{all clusters}}P(i\in C) \cdot \log{P(i\in C)}
\end{align}
Where $P(i\in C)$ is defined as in Equation~\ref{eq:clustered_probabilities}.
%Note that if we didn't use clustered probabilities, we would get the formula $-\sum_{y\in\mathcal{Y}(i)}P(\text{ant}(i) = y) * \log{P(\text{ant}(i) = y)}$, which is quite similar to the global entropy $GE1$ formula from \newcite{Gasperin:2009:ALA:1564131.1564133} (only she uses multi-class entropy)

\paragraph{Clustered query-by-committee} We train $\mathcal{M}$ models (with different random seeds) and select the mention with the highest \emph{cluster} vote entropy:
\begin{align}
    \text{VE}(i) = -\sum_{C\in\text{all clusters})}\frac{V(i\in C)}{\mathcal{M}} \cdot \log{\frac{V(i\in C)}{\mathcal{M}}}
\end{align}
Using votes counted over clusters, as defined in Equation~\ref{eq:clustered_votes}.%\footnote{The modified training loop we use for the query-by-committee selector is shown in Algorithm \ref{alg:train_qbc} in the Appendix.}

\paragraph{Least coreferent clustered mentions / Most coreferent unclustered mentions (LCC/MCU)} 
We aim to select a subset of spans for which the model was least confident in its prediction. 
For each span $i$ which was assigned a cluster $C_i$, we compute a score $s_C(i) = P(i\in C_i)$, and choose $n$ spans with the smallest $s_C(i)$. For each singleton $j$, we give an ``unclustered'' score $s_U(i) = \max_{C\in \text{all clusters}}P(j\in C)$ and choose $m$ spans with the \emph{largest} $s_U(i)$. $P(i\in C_i)$ and $P(j\in C)$ are computed with Equation \ref{eq:clustered_probabilities}.

% \paragraph{Pairwise Entropy (Pairwise Baseline).} To compare our approach with pairwise annotation, we implement an entropy selector for pairwise annotation. We select the next span pair $(i, a)$ to annotate based on entropy:
% $$E(i, a) = -P(\text{ant}(i) = a) * \log{P(\text{ant}(i) = a)}$$
% where $a\in\mathcal{Y}(i)$. To ensure we don't select the same pair twice, we set $P(\text{ant}(i) = a)$ to $0$ if the pair was deemed not coreferent, and to $1$ if the pair was deemed coreferent.

\section{Evaluation}
\label{sec:evaluation}
% In this section we compare discrete annotation with pairwise annotation.
%\subsection{Simulating experiments}
We compare discrete versus pairwise annotation using the English CoNLL-2012 coreference dataset \cite{pradhan2012conll}. Following \newcite{Sachan:2015:ALA:2832415.2832432}, we conduct experiments where user judgments are simulated from gold labels.
% For each document, we also keep track of the number of positive and negative user answers to the initial question to best estimate annotation time.

\begin{figure}
    \centering
    \includegraphics[width=\columnwidth,keepaspectratio]{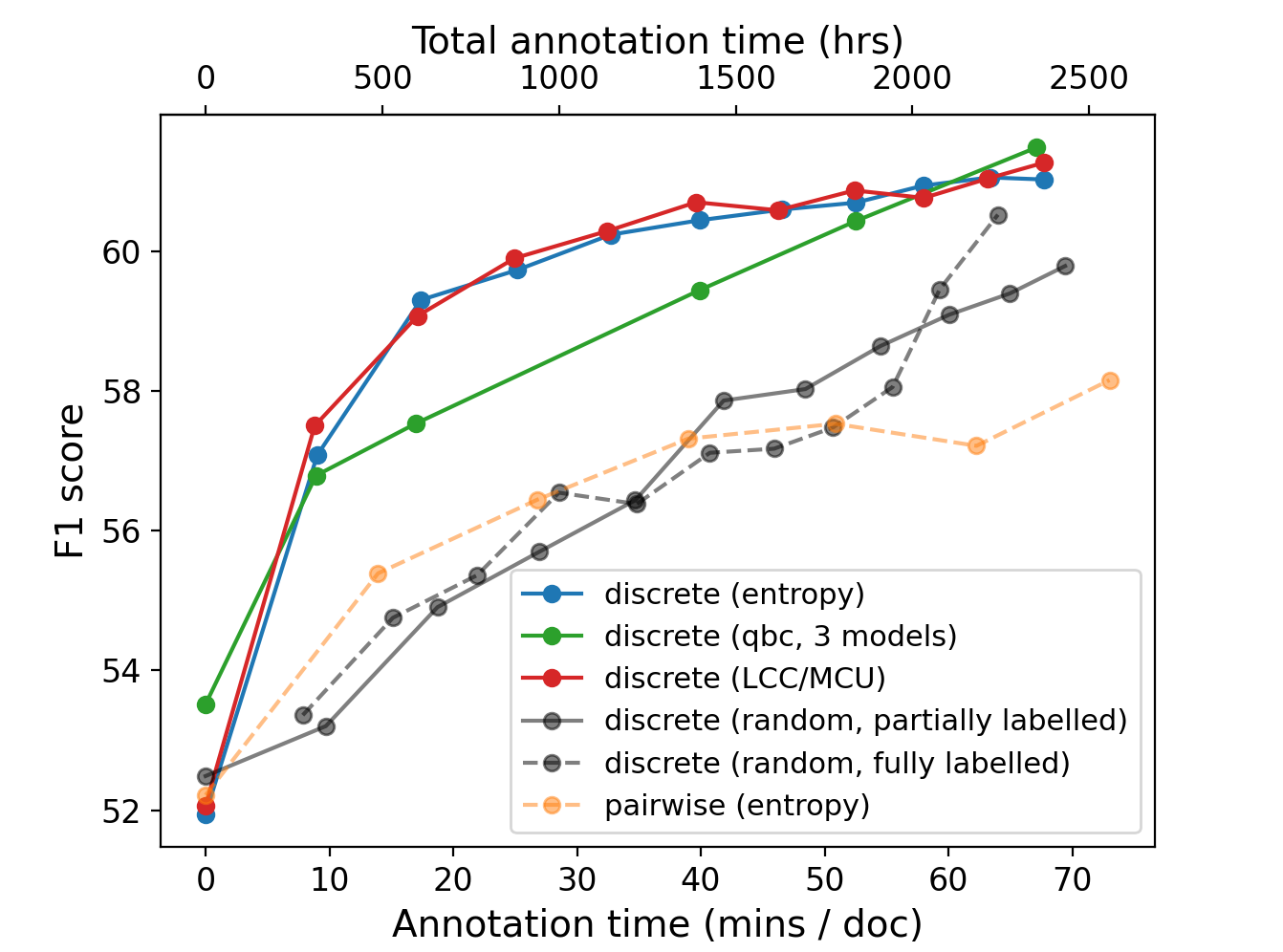}
    \caption{Comparing various selectors for discrete  versus pairwise annotation (dashed orange line).}
    \label{fig:selectors}
\end{figure}

\begin{table}[tb!]
\small
\centering
\begin{tabular}{ccccc} \hline
 & & Active learning & & \\
Set & \# labels/doc & iteration & \# docs & \# ?s \\ \hline \hline
& 20 & 1st (retrained 0x) & 5 & 15 \\
A & 20 & 7th (retrained 6x) & 5 & 15 \\
& 200 & 2nd (retrained 1x) & 5 & 15 \\
& 200 & 8th (retrained 7x) & 5 & 15 \\ \hline
& 20 & 2nd (retrained 1x) & 5 & 15 \\
B & 20 & 8th (retrained 7x) & 5 & 15 \\
& 200 & 1st (retrained 0x) & 5 & 15 \\
& 200 & 7th (retrained 6x) & 5 & 15 \\ \hline
\end{tabular}
\caption{Timing experiments sampling. For each of the 2 datasets, we collected 60 total active learning questions from 20 documents. We collected 5 documents and 15 questions for each of the 4 categories: trained with many/few labels per document, and early/late in active learning process. The 15 questions were sampled randomly from within an iteration.}
\label{tab:timing_sampling}
\end{table}

\paragraph{Annotation time estimation}
%\label{sec:time}
% Since the motivation for active learning is optimizing annotation budget (e.g., in terms of monetary and time investment), it is vital that we estimate the time it takes to perform discrete annotation vs. pairwise. % Doing so will allow us to compare the two approaches on an equal footing. 
To compare annotation times between pairwise and discrete questions, we collected eight 30-minute sessions from 7 in-house annotators with background in NLP. Annotators were asked to answer as many instances as they could during those 30 minutes. We additionally asked 1 annotator to annotate \textit{only} discrete questions for 30 minutes. 
To be as representative as possible, the active learning queries for these experiments were sampled from various stages of active learning (see Table~\ref{tab:timing_sampling}).
On average, an annotator completed about 67 questions in a single session, half of which were answered negatively, requiring the additional discrete question. Overall, these estimates rely on 826 annotated answers.
Our annotation interface is publicly available,\footnote{\url{https://belindal.github.io/timing_experiments}} see examples in Figure~\ref{fig:interface} in the Appendix.

Timing results are shown in Table~\ref{Tab:time_results}.
Answering the discrete question after the initial pairwise question takes about the same time as answering the first question (about $16s$).
Furthermore, answering only discrete questions took $28.01s$ per question, which confirmed that having an initial pairwise question indeed saves annotator time if answered positively.

In the following experiments, we use these measurements 
to calibrate pairwise and discrete followup questions when computing total annotation times.

\paragraph{Baselines}
We implement a baseline for pairwise annotation with entropy selector. We also implement two discrete annotation baselines with random selection. The \emph{partially-labelled} baseline follows the standard active learning training loop, but selects the next mention to label at random. The \emph{fully-labelled} baseline creates a subset of the training data by taking as input an annotation time $t$ and selecting at random a set of documents that the user can \textit{fully} label in $t$ hours using ONLY discrete annotation. By comparing the fully-labelled baseline against our active learning results, we can determine whether active learning is effective over labelling documents exhaustively .

\paragraph{Hyperparameters}
We use the model hyperparameters from the AllenNLP implementation of \newcite{Lee2017EndtoendNC}. 
We train up to 20 epochs with a patience of 2 before  adding labels. After all documents have been added, we retrain from scratch.
We use a query-by-committee of $\mathcal{M} = 3$ models, due to memory constraints.
For LCC/MCU, given $L$ annotations per document, we split the annotations equally between clusters and singletons.

%, showing that this additional annotation time is indeed cost-effective.

\begin{table}
%\small
\centering
\begin{tabular}{lr} \hline
     & Avg. Time per ? \\ \hline \hline
    %Discrete instances (overall) & $24.10s$ \\
    Initial question & $15.96s$ \\% \pm 13.91s$ \\
    Follow-up question & $15.57s$ \\% \pm 17.39s$ \\
    %Initial + follow-up question & $35.05s$ \\
    ONLY Follow-up questions & $28.01s$ \\
    \hline
\end{tabular}
\caption{Average annotation time for the initial pairwise question, the discrete followup question, and the discrete question on its own.}
\label{Tab:time_results}
\end{table}

\paragraph{Results}
%We compare active learning methods (discrete vs. pairwise, various selection techniques) by plotting F1 score against annotation time
%Figure~\ref{pairwise_vs_discrete} plots pairwise vs. discrete annotation, with both using an entropy selector. With the same annotation time spent per document, it is clear that discrete outperforms pairwise.
Figure~\ref{fig:selectors} plots the performance of discrete annotation with the various selectors from Section \ref{sec:selectors}, against the performance of pairwise annotation, calibrated according to our timing experiments. In all figures, we report MUC, B3, and CEAFe as an averaged F1 score.
% Note the $x$-axis is annotation time, which we compute using the formulas from our timing experiments (Equations \ref{eq:timing_pairwise}, \ref{eq:timing_discrete}, \ref{eq:timing}).

The three non-random active learning frameworks outperform the fully-labelled baseline, showing that active learning is  more effective for coreference resolution when annotation budget is limited.

Most notably, Figure~\ref{fig:selectors} shows that every non-random discrete selection protocol outperforms pairwise annotation.
Where the gap in performance is the largest ($>15$ minutes per document), we consistently improve by $\sim$4\% absolute $F1$ over pairwise selection. %For reference, this is comparable to the gains reported by \citet{Sachan:2015:ALA:2832415.2832432}, when normalized by annotation time.
%Moreover, it appears that the entropy and LCC/MCU selectors outperform query-by-committee for discrete annotation despite query-by-committee using an ensemble model.

\section{Analysis}
A major reason discrete annotation outperforms the pairwise baseline is that the number of pairwise annotations needed to fully label a document is much larger than the number of discrete annotations. In an average development document with $201$ candidates per mention,
the number of pairwise queries needed to fully label a document is $15,050$, while the maximum number of discrete queries is only $201$ (i.e., asking for the antecedent of every mention).
Thus, the average document can be fully annotated via discrete annotation in only 2.6\% of the time it takes to fully label it with pairwise annotation, suggesting that our framework is also a viable \emph{exhaustive} annotation scheme.

% Moreover, we find that our method saves time over simply asking the follow-up question. For example, to get to $60\%$ F1, the only-follow-up baseline needed $1455h$ annotation hours, while the full annotation scheme took $1200h$.
%For the entropy selector, around $23.7\%$ of initial questions were answered positively. From our timing experiments, we compute that maintaining the initial question saved $256h$ on average over the $10$ different annotation times we measured.

%, perhaps because of the low number of models in our committee. Unfortunately, limitations on machine memory and training time prevent us from using more models.
% \begin{figure}
%     \centering
%     \includegraphics[scale=0.5]{images/pairwise_vs_discrete_entropy.png}
%     \caption{Comparing discrete and pairwise annotation, with an entropy selector. $x$-axis normalized by annotation time}
%     \label{pairwise_vs_discrete}
% \end{figure}

\begin{figure}
    \centering
    \includegraphics[width=\columnwidth,keepaspectratio]{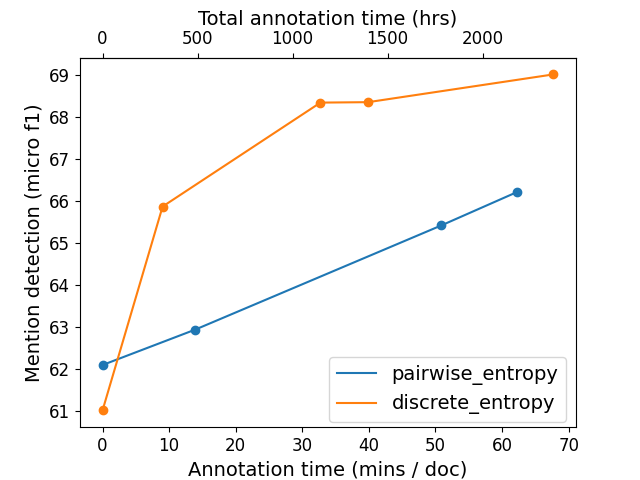}
    \caption{Mention detection accuracy (in document-micro F1) for pairwise versus discrete selection per human annotation time.}
    \label{fig:mention}
\end{figure}

\begin{figure}
    \centering
    \includegraphics[width=\columnwidth,keepaspectratio]{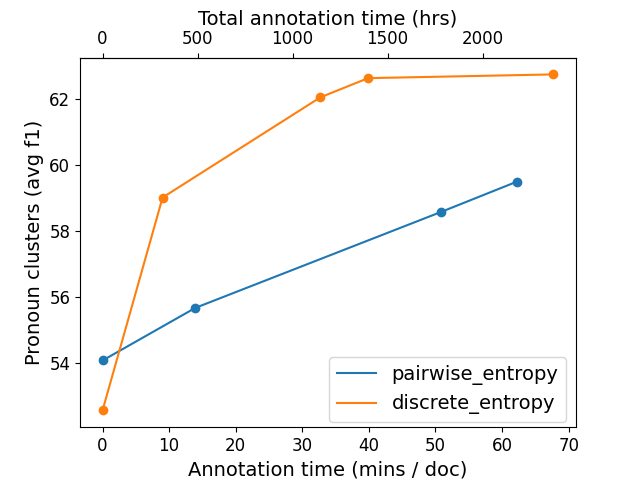}
    \caption{Pronoun resolution accuracy (average F1) for pairwise versus discrete selection per human annotation time.}
    \label{fig:pronoun}
\end{figure}

\label{sec:analysis}
%how much using clustered selectors and recomputing link closures after each annotation contribute to the final F1 score.

Further analysis shows that the improvement in  discrete selection stems in part from better use of annotation time for mention detection accuracy (Figure~\ref{fig:mention}) and pronoun resolution (Figure~\ref{fig:pronoun}), in which we measure performance only on clusters with pronouns, as identified automatically by the spaCy tagger \cite{spacy2} .

Finally, Table~\ref{tab:ablation} shows ablations on our discrete annotation framework, showing the contribution of each component of our paradigm.

\begin{table}
    \centering
    %\small
    \begin{tabular}{lc} \hline
        & F1 score \\ \hline \hline
        Discrete annotation  & 57.08 \\
        $-$clustered probabilities  & 56.49 \\
        $-$incremental link  & 56.98 \\
        \,\,\, closures & \\ \hline
        Pairwise annotation & 54.27 \\ \hline 
    \end{tabular}
    \caption{Ablations over the different model elements, at a single point ($\sim$315 annotation hours). Entropy selector was used for all experiments.}
    \label{tab:ablation}
\end{table}

\section{Discussion and Conclusion}
We presented discrete annotation, an attractive alternative to pairwise annotation in active learning of coreference resolution in low-resource domains. By adding a simple question to the annotation interface, we obtained significantly better models per human-annotation hour.
In addition, we introduced a clustering technique which further optimizes sample selection during the annotation process. 
More broadly, our work suggests that improvements in annotation interfaces can elicit responses which are more efficient in terms of the obtained performance versus the invested annotation time.

\section*{Acknowledgements}
We would like to thank Christopher Clark, Terra Blevins, and the anonymous reviewers for their helpful feedback, and 
Aaron Jaech, Mason Kamb, Madian Khabsa, Kaushal Mangipudi, Nayeon Lee, and Anisha Uppugonduri for their participation in our timing experiments.
%\clearpage

% \section{Use Case: Annotating coreference in a new domain}
% \input{use_case}

\bibliography{acl2020}
\bibliographystyle{acl_natbib}

\newpage
\appendix
\section{Appendix}

\subsection{Timing Experiment Details and Computations.}
\label{sec:timing_experiment_appendix}
\begin{figure*}[ht]
    \centering
    \fbox{\includegraphics[scale=0.48]{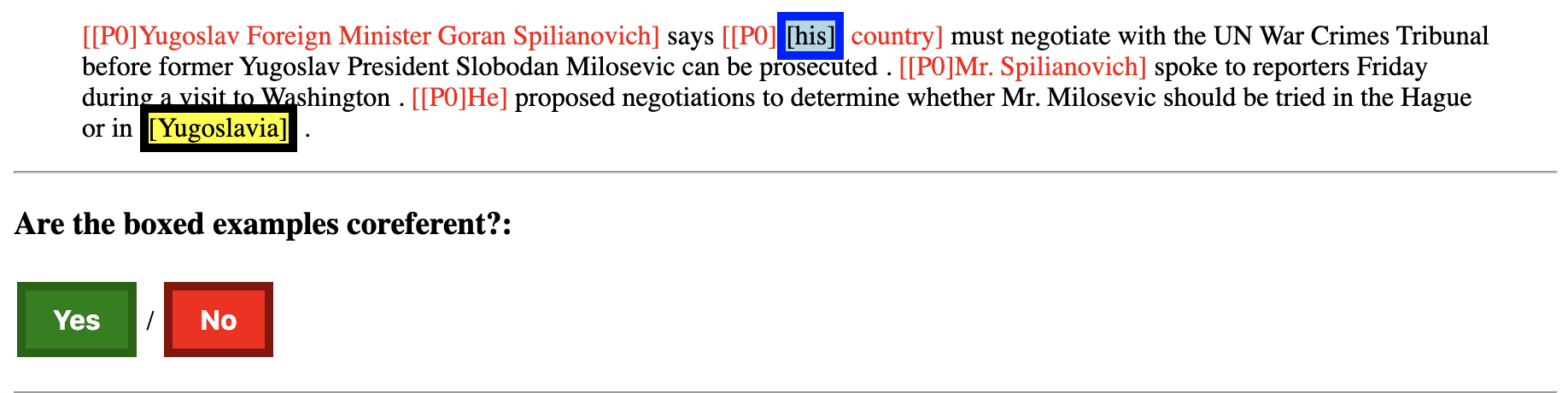}}
    %\fbox{\includegraphics[scale=0.5]{images/interface/interface_discrete.png}}
    \fbox{\includegraphics[scale=0.48]{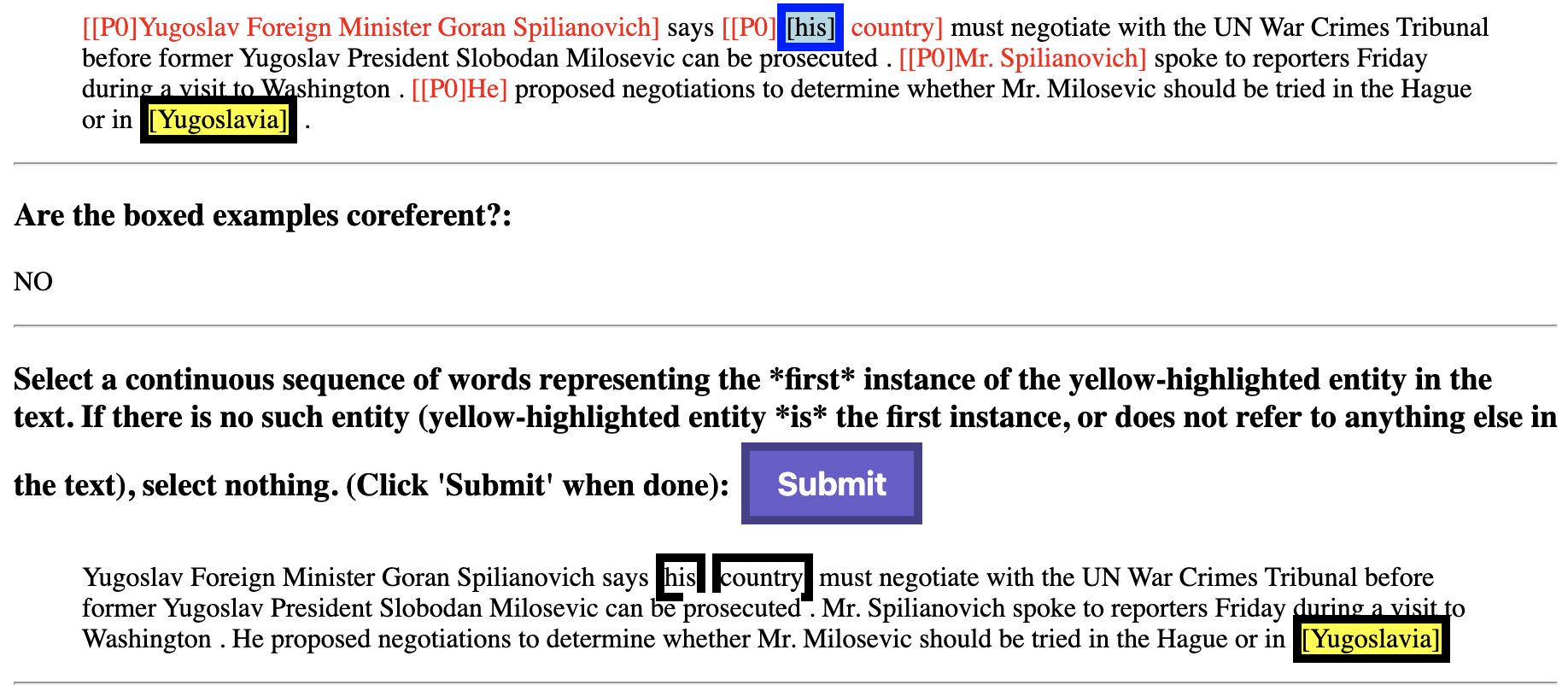}}
    \caption{Timing experiments interface. Top: The initial pairwise question. Bottom: The user is presented with the discrete question when they click ``No". They are asked to select the appropriate tokens in the text representing the first occurrence of the yellow entity in the text.}
    \label{fig:interface}
\end{figure*}

In order to properly calibrate the results from discrete and pairwise querying, we conducted experiments (eight 30-minute sessions) to time how long annotators take to answer discrete and pairwise questions. See Figure~\ref{fig:interface} for the interface we designed for our experiments.

The questions we ask for the experiment are all sampled from real queries from full runs of our active learning simulations. To obtain representative times, we sampled a diverse selection of active learning questions--at various stages of active learning (first iteration before retraining vs. after retraining $n$ times) and various numbers of annotation per document (20 vs. 200). For each document, we randomly selected between 1-5 questions (of the total 20 or 200) to ask the annotator. Full details on how we sampled our queries can be found in Table~\ref{tab:timing_sampling}. Note that we divided our samples into two datasets. We ran four 30-minute sessions with Dataset A before Dataset B and four 30-minute sessions with Dataset B before Dataset A--for a total of eight 30-minute sessions across 7 annotators (1 annotator completed a 1-hour session).

Since pairwise annotation is the same as answering only the initial question under the discrete setting, we run a single discrete experiment for each annotation session and use the time taken to answer an initial question as a proxy for pairwise annotation time.
Our results show that answering the initial question took an average of $15.96s$ whereas answering the follow-up question took $15.57s$. Thus, we derive the following formulas to compute the time it takes for pairwise and discrete annotation:
\begin{align}
    t &= 15.96p \label{eq:timing_pairwise} \\
    t &= 15.96d_c + 15.57d_{nc} \label{eq:timing_discrete}
\end{align}
where $p = \#$ of pairwise instances. $d_c, d_{nc} = \#$ of discrete instances for which the initial pair was ``coreferent" ($d_c$) and ``not coreferent" ($d_{nc}$), respectively. We also compute the number of pairwise examples $p$ we can query in the same time it takes to query $d_c+d_{nc}$ discrete examples:
\begin{align}
15.96p & = 15.96d_c + 15.57d_{nc} \nonumber \\
p & = d_c + 0.976d_{nc} \label{eq:timing}
\end{align}

Moreover, we additionally conduct a single 30-minute experiment to determine how long it takes to answer \textit{only} discrete questions (without the initial pairwise step). We find that it takes $28.01s$ per question under the only-discrete setting. This is longer than the time it takes to answer a pairwise question, thus confirming that having an initial pairwise question indeed saves time if the pair is coreferent. Moreover, this also shows that  answering the initial pairwise question significantly helps with answering the follow-up discrete question.

\subsection{Additional Model Adaptations}
\label{sec:model_appendix}

\paragraph{Adapting Link Relations for our Model}
We use must-link and cannot-link relations between mentions to guide our active learning selector. We revise probabilities and model outputs (from which the model computes uncertainty scores for entropy, QBC, and LCC/MCU) in accordance to the following rules:

\begin{enumerate}
\item \textbf{Clustered entropy}. For every $CL(a,b)$ relationship, we set $P(\text{ant}(a) = b) = 0$ and re-normalize probabilities of all other candidate antecedents. This decreases the probability that the active learning selector chooses $a$. Moreover, for every $ML(a,b)$ relationship, we set $P(\text{ant}(a) = b) = 1$ and $P(\text{ant}(a) = c) = 0$ for all $c\neq b$. If there are multiple $ML$ relationships involving $a$, we choose only one of $a$'s antecedent to set to $1$ (to maintain the integrity of the probability distribution). This guarantees that the active learning selector will never select $a$, as any ML link out of $a$ means we have already queried for $a$.

\item \textbf{Clustered query-by-committee}. To ensure we do not choose a mention we have already queried for, after each user judgment, for every $ML(a,b)$ relation, we set $V(A(a) = b) = \mathcal{M}$, and $V(A(a) = c) = 0$ for all other $c\neq b$. Moreover, for every $CL(a,b)$ relation, we set $V(A(a) = b) = 0$, which decreases the vote entropy of $a$, making it less likely for the selector to choose $a$.

\item \textbf{LCC/MCU}. We revise the probabilities in the same way as in clustered entropy and add the constraint that, when choosing MCU spans $j$, we disregard those that already have probability $1$ (signifying that we have already queried for them).
\end{enumerate}

\paragraph{Incremental Closures Algorithm}
%\label{sec:analysis_appendix_inc}
We introduce an algorithm to compute link closures
%% TODO link closures here
\emph{incrementally}. Instead of re-computing and re-adding the entire set of closures (based on a set of all prior human annotations that we keep track of) each time we query for a new mention, we add the minimum set of necessary links. See Algorithm~\ref{alg:inc_closures}. 

To determine how much time our incremental closure algorithm saves over recomputing closures from scratch, we simulated annotations on a single document with $1600$ mentions, and recorded how long it took to re-compute the closure after each annotation. Our experiments show that recomputing from scratch takes progressively longer as more labels get added: at $1600$ labels, our incremental algorithm is $556$ times faster than recomputing from scratch ($1630ms$ vs. $2.93ms$).

\begin{figure}
    \centering
    \includegraphics[scale=0.5]{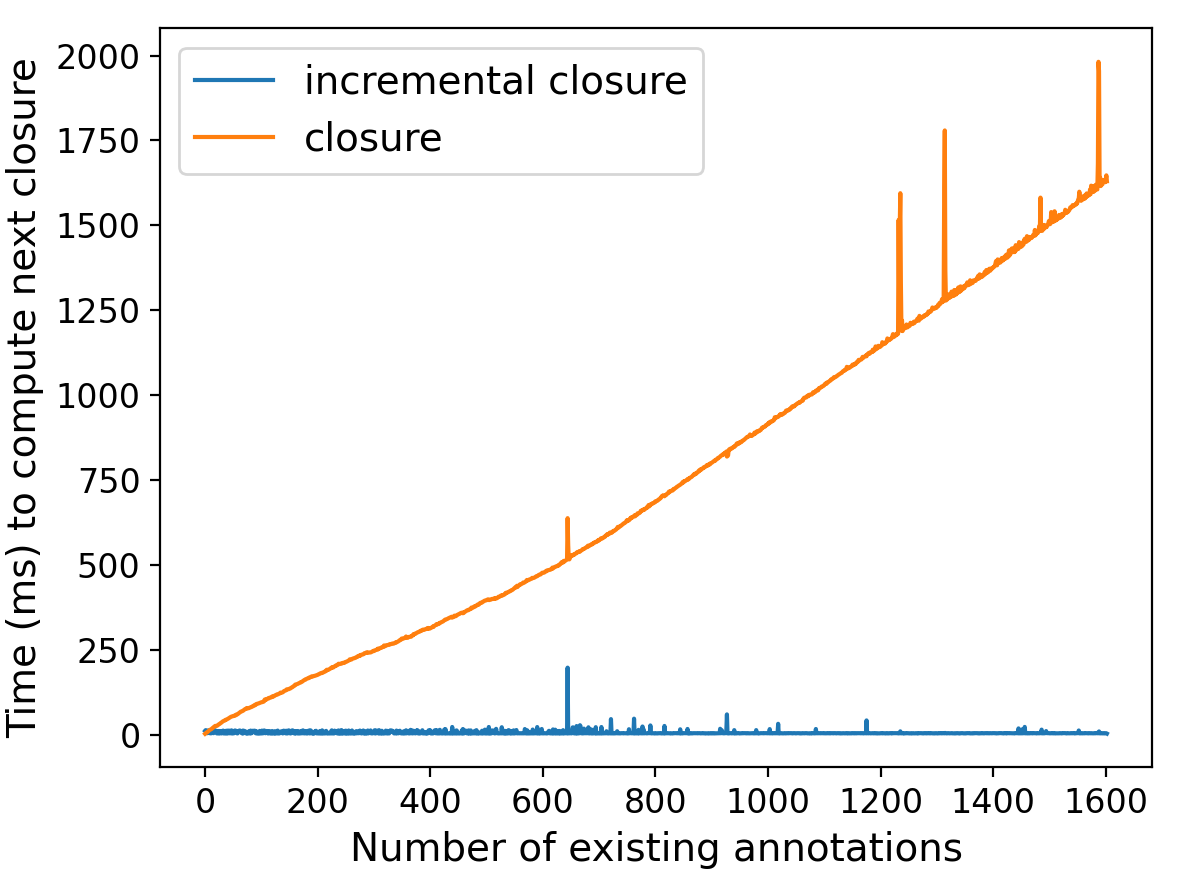}
    \caption{Under each closure algorithm, the time to compute the closure after the next annotation is added, as \# of existing annotations increases.}
    \label{fig:closure_runtimes}
\end{figure}

Figure~\ref{fig:closure_runtimes} plots the runtime of our incremental closure algorithm (``incremental closure") against the run-time of recomputing closures from scratch (``closure") using Equations~\ref{eq:closure1} and \ref{eq:closure2}. In the latter case, we keep track of the set of user-added edges which we update after each annotation, and re-compute the closures from that set.

\subsection{Additional Analysis}
\label{sec:analysis_appendix}

\paragraph{Computing the time to fully-label a document under discrete and pairwise annotation.}
First, we compute the maximum number of pairwise questions we can ask. We consider the setup of \newcite{Lee2017EndtoendNC}'s model. This model considers only spans with highest mention scores (the ``top spans"), and only considers at most $K$ antecedents per top span. Thus, for a document with $m$ top spans, we can ask up to
\begin{align}\label{eq:max_pairwise_questions}
\frac{K(K-1)}{2} + (m - K)K
\end{align}
pairwise questions. The first factor $\frac{K(K-1)}{2}$ comes from considering the first $K$ spans in the document. For each of these spans $i=1\cdots K$, we can ask about the first $i-1$ spans. The second factor $(m - K)K$ comes from considering the spans after the $K$-th span. For each of these $m - K$ spans in the document, we can only consider up to $K$ antecedents. Using statistics for the average document ($m=201$) and the standard hyper-parameter settings ($K=100$), we plug into Equation~\ref{eq:max_pairwise_questions} to get $15,050$ overall pairwise questions needed to fully label a document (in worst-case). Meanwhile, the maximum number of discrete questions we can ask is only $201$ (i.e., asking for the antecedent of every mention). Using timing Equations \ref{eq:timing_pairwise} and \ref{eq:timing_discrete}, we compute that it takes at most $6337.53s$ to answer $201$ discrete questions in the worst-case scenario, and $240198s$ to answer $15050$ pairwise questions. Thus, in the worst-case scenario for both discrete and pairwise selection, discrete selection will take only $2.64\%$ of the time it takes pairwise selection to fully label a document.

\paragraph{Quantifying ``Information Gain" from Discrete and Pairwise Annotation.}
Let $\overline{D_U}$ be the set of training documents we are \textit{annotating for} in a given round of active learning. To better quantify how much information discrete and pairwise annotation can supply in same amount of time, we define $\Delta F1$ as the change in the $F1$ score on $\overline{D_U}$, before and after model predictions are supplemented with user annotation. 
% %, where $F1_0$ is the original model F1 score on $\overline{D_U}$, and $F1_1$ is the F1 score on $\overline{D_U}$ after the model predictions are supplemented with user annotation.

\begin{figure}
    \centering
    \includegraphics[scale=0.5]{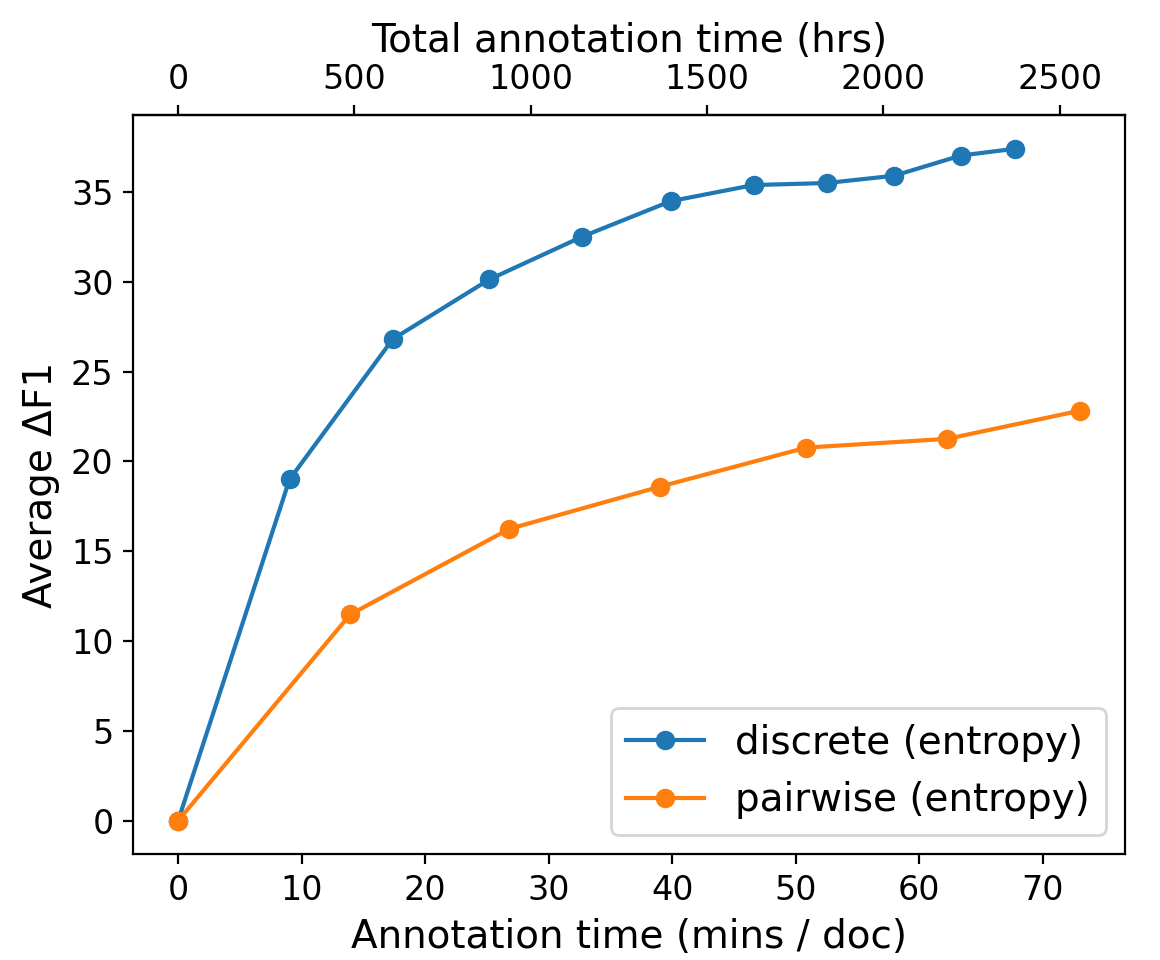}
    \caption{Comparing F1 score improvement on $\overline{D_U}$ for discrete vs. pairwise annotation.}
    \label{fig:deltas}
\end{figure}

Figure~\ref{fig:deltas} shows average $\Delta F1$ as annotation time increases for discrete and pairwise annotation. 
Across the 10 annotation times we recorded, discrete annotation results in an average $\Delta F1$ that more than twice that of pairwise, in the same annotation time.

\subsection{Hyperparameters}
\paragraph{Model.} We preserve the hyperparameters from the AllenNLP implementation of \newcite{Lee2017EndtoendNC}'s model. The AllenNLP implementation mostly maintains the original hyperparameters, except it sets the maximum number of antecedents considered to $K = 100$, and excludes speaker features and variational dropout, due to machine memory limitations. %The AllenNLP implemention is able to achieve an F1-score of $63\%$ on the fully-labelled dataset.
%As shown in Figures \ref{pairwise_vs_discrete} and \ref{discrete_selectors}, we are able to obtain close to such a score, with much less labels, by using active learning with discrete selection.

\paragraph{Training.} We use a 700/2102 fully-labelled/unlabelled initial split of the training data, and actively label 280 documents at a time. We train to convergence each round. Before all documents have been added, we train up to 20 epochs with a patience of 2 before we add more training documents. After all documents have been added, we retrain from scratch and use the original training hyperparameters from \newcite{Lee2017EndtoendNC}.

\paragraph{Selectors.} For query-by-committee, we use a committee of $\mathcal{M} = 3$ models. We were not able to experiment with more due to memory constraints. 

For LCC/MCU, given $L$ annotations per document, we allocate $n$ annotations to least-coreferent clustered mentions and the remaining $m$ to most-coreferent unclustered mentions. We use $n = \min{(L/2, \text{number of clustered spans})}$, and \\
$m = \min(L - n,$ number of un-clustered spans$)$.

\subsection{Active Learning Training Setup Full Details}
In our active learning setup, we begin by training our model on a 700-document subset of the full training set. We discard the labels of the remaining 2102 documents. %Let $D_F$ be the fully-labelled set and $D_U$ be the unlabelled set.
In each round of active learning, we choose 280 unlabelled documents, and query up to $Q$ annotations per document.
We then add these documents to the labelled set and continue training our model on this set (now with new documents).
%through active learning, and add them to a set $D_A$.
After all documents have been labelled, we retrain our model on the full document set from scratch, resetting all model and trainer parameters.

In Algorithm~\ref{alg:train}, we show our main training loop for active learning using discrete selection. This is the training loop we use for our clustered entropy and LCC/MCU selectors, and our partially-labelled random baseline. In Algorithm~\ref{alg:train_qbc}, we modify that loop for the clustered query-by-committee selector.

In Algorithm~\ref{alg:inc_closures}, we show our incremental closures algorithm, which builds up the transitive closure incrementally by adding only the minimum number of necessary links to maintain the closure each time a new link is added.

% \begin{algorithm*}
% \caption{Incremental Link Closures Algorithm}
% \label{alg:inc_closures}
% \begin{algorithmic}[1]
% \small
% \STATE {$(a,b) = $ link pair being added (to either must-link or cannot-link) }
% \IF{$(a,b)$ was added to must-link}
%     \STATE { $\forall a',b'\quad (ML(a,a')\land ML(b,b'))\to (ML(a,b')\land ML(a',b)\land ML(a',b'))$ } \COMMENT{ first resolve $ML$s }
%     % \STATE{ $\forall a',\hat{b}\quad (ML(a,a')\land CL(b,\hat{b}))\to (CL(a,\hat{b})\land CL(a',\hat{b}))$ } \COMMENT{ then resolve $CL$s (using resolved $ML$s) }
%     \STATE{ $\forall b',\hat{a}\quad (ML(b,b')\land CL(a,\hat{a}))\to (CL(b,\hat{a})\land CL(b',\hat{a}))$ } \COMMENT{ resolve $CL$s }
% \ELSE
%     \STATE { $\forall a', b'\quad (ML(a,a')\land ML(b,b'))\to (CL(a,b')\land CL(a',b)\land CL(a',b'))$ } \COMMENT{ resolve $CL$s }
% \ENDIF
% \end{algorithmic}
% \end{algorithm*}
\begin{algorithm*}[b!]
\small
Let $(a,b) = $ link pair being added, $A = a$'s old cluster before the pair is added, $B = b$'s old cluster before the pair is added, $\overline{A} = $ set of element $a$ has a CL relationship to before the pair is added, $\overline{B} = $ set of elements $b$ has a CL relationship to before the pair is added.
\begin{enumerate}
    \item If pair $(a,b)$ was added to \textit{must-link}, both must-link and cannot-link needs to be updated. \\
    First, resolve the MLs by adding a ML relationship between every element in $A$ and every element in $B$:
    $$\forall a',b'\quad (ML(a,a')\land ML(b,b'))\to (ML(a,b')\land ML(a',b)\land ML(a',b'))$$
    Next, resolve the CLs by adding a CL relationship between every element of $A$ and $\overline{B}$, and every element of $B$ and $\overline{A}$:
    $$\forall a',\hat{b}\quad (ML(a,a')\land CL(b,\hat{b}))\to (CL(a,\hat{b})\land CL(a',\hat{b}))$$
    $$\forall b',\hat{a}\quad (ML(b,b')\land CL(a,\hat{a}))\to (CL(b,\hat{a})\land CL(b',\hat{a}))$$
    \item If pair $(a,b)$ was added to \textit{cannot-link}, only cannot-link needs to be updated. Add a CL relationship between every element of $A$ and every element of $B$:
    $$\forall a', b'\quad (ML(a,a')\land ML(b,b'))\to (CL(a,b')\land CL(a',b)\land CL(a',b'))$$
\end{enumerate}
\caption{Incremental Link Closures Algorithm}
\label{alg:inc_closures}
\end{algorithm*}

\begin{algorithm*}
\small
$D_F$ = \{fully-labelled docs\}, $D_U$ = \{unlabelled docs\}, $D_A$ = \{docs labelled through active learning\}, $M$ = model, $ML$ = must-link pairs, $CL$ = cannot-link pairs; \\
Init: $D_F$ = \{first 700 docs\}, $D_U$ = \{remaining docs\}, $D_A = \emptyset$, $ML = CL = \emptyset$; \\
\While{$D_U$ is not empty}{
    train $M$ to convergence on data $D_F\cup D_A$; \\
    $\overline{D_U}$ = 280-document subset of $D_U$; \\
    \For{$D\in \overline{D_U}$}{
        $\mathcal{P}_D, \mathcal{L}_D, \mathcal{C}_D$ = run $M$ on $D$; \\
            \quad $\mathcal{P}_D$ = model-outputted probabilities = $\{P(y = \text{ant}(i)) | y\in\mathcal{Y}(i), i\in\text{top\_spans}(D)\}$ \\
            \quad $\mathcal{L}_D$ = model-outputted antecedent labels = $\{(i, A(i)) | i\in\text{top\_spans}(D)\}$ \\
            \quad $\mathcal{C}_D$ = model-outputted clusters from $\mathcal{L}_D$ \\
        \While{num\_queried $<$ num\_to\_query}{
            $m$ = choose-next-mention-to-query($\mathcal{P}_D, \mathcal{C}_D$);\qquad\qquad [[Section~\ref{sec:selectors}]]\\
            $a$ = $\max_{y\in\mathcal{Y}(m)\backslash\epsilon}P(y = \text{ant}(m)) $; \\
            \uIf{user deems $m$ and $a$ coreferent}{
                $ML = ML\cup (a, m)$; \\
                $\mathcal{L}_D = \mathcal{L}_D\cup (a,m)$; \\
                Add $(a,m)$ to $\mathcal{C}_D$; \\
            }\Else{
                $\hat{a}$ = user-selected antecedent for $m$; \\
                $CL = CL\cup (a, m)$; $ML = ML\cup (\hat{a}, m)$; \\
                $\mathcal{L}_D = (\mathcal{L}_D \backslash (a,m)) \cup (\hat{a},m)$; \\
                Remove $(a,m)$ and add $(\hat{a},m)$ to $\mathcal{C}_D$; \\
            }
            $ML, CL$ = compute-link-closures;\qquad\qquad\qquad\qquad [[Algorithm~\ref{alg:inc_closures}]]\\
            $\mathcal{P}_D$ = update-based-on-links($ML$, $CL$);\qquad\qquad\qquad[[Section~\ref{sec:model_appendix}]]\\
        }
        Label $D$ with $\mathcal{C}_D$; \\
    }
    $D_A = D_A\cup \overline{D_U}$; $D_U = D_U \backslash \overline{D_U}$; \\
}
\caption{Training loop for active learning}
\label{alg:train}
\end{algorithm*}

\begin{algorithm*}
\small
$D_F$ = \{fully-labelled docs\}, $D_U$ = \{unlabelled docs\}, $D_A$ = \{docs labelled through active learning\}, $\widehat{M} = $ ensemble model of submodels $\{M_1,\cdots, M_\mathcal{M}\}$, $ML$ = must-link pairs, $CL$ = cannot-link pairs; \\
Init: $D_F$ = \{first 700 docs\}, $D_U$ = \{remaining docs\}, $D_A = \emptyset$, $ML = CL = \emptyset$; \\
\While{$D_U$ is not empty}{
    \HiLi train all $M_1,\cdots, M_\mathcal{M}$ to convergence on data $D_F\cup D_A$; \\
    $\overline{D_U}$ = 280-document subset of $D_U$; \\
    \For{$D\in \overline{D_U}$}{
        \HiLi $\{\mathcal{P}_{D,i}\}, \{\mathcal{L}_{D,i}\}, \mathcal{P}_{D}, \mathcal{L}_{D}, \mathcal{C}_{D}$ = run $\widehat{M}$ on $D$; \\
            \HiLi \quad $\mathcal{P}_{D,i}$ = submodel $i$'s output probabilities \\
            \HiLi \quad $\mathcal{L}_{D,i}$ = submodel $i$'s output antecedent labels \\
            \HiLi \quad $\mathcal{P}_{D}$ = ensembled (averaged) output probabilities from each submodel \\
            \HiLi \quad $\mathcal{L}_{D}$ = ensembled antecedent labels computed from $\mathcal{P}_{D}$ \\
            \HiLi \quad $\mathcal{C}_{D}$ = ensembled clusters computed from $\mathcal{L}_{D}$ \\
        \While{num\_queried $<$ num\_to\_query}{
            \HiLi $m$ = choose-next-mention-to-query($\{\mathcal{L}_{D,i}\}, \mathcal{C}_{D}$);\qquad\,\,[[Section~\ref{sec:selectors}]]\\
            $a$ = $\max_{y\in\mathcal{Y}(m)\backslash\epsilon}P(y = \text{ant}(m)) $;\qquad\qquad\qquad\qquad\,[[Probabilities from $\mathcal{P}_D$]] \\
            \uIf{user deems $m$ and $a$ coreferent}{
                $ML = ML\cup (a, m)$; \\
                Add $(a,m)$ to $\mathcal{C}_D$; \\
            }\Else{
                $\hat{a}$ = user-selected antecedent for $m$; \\
                $CL = CL\cup (a, m)$; $ML = ML\cup (\hat{a}, m)$; \\
                Remove $(a,m)$ and add $(\hat{a},m)$ to $\mathcal{C}_D$; \\
            }
            $ML, CL$ = compute-link-closures($ML,CL$);\qquad\quad\,\, [[Algorithm~\ref{alg:inc_closures}]]\\
            \HiLi $\mathcal{L}_{D,i}$ = update-based-on-links($ML$, $CL$); \qquad\qquad\,\quad [[Section~\ref{sec:model_appendix}]]\\
        }
        Label $D$ with $\mathcal{C}_D$; \\
    }
    $D_A = D_A\cup \overline{D_U}$; $D_U = D_U \backslash \overline{D_U}$; \\
}
\caption{Training loop for active learning with QBC selector (Differences from Algorithm~\ref{alg:train} are highlighted)}
\label{alg:train_qbc}
\end{algorithm*}

\end{document}